\documentclass[runningheads]{llncs}
\usepackage{graphicx}

\usepackage{tikz}
\usepackage{comment}
\usepackage{amsmath,amssymb} 
\usepackage{color}

\usepackage{algorithm}
\usepackage{algpseudocode}
\usepackage{booktabs}
\usepackage{caption}
\usepackage{subcaption}
\usepackage{multirow}
\usepackage{makecell}
\usepackage{hyperref}

\begin{document}
\pagestyle{headings}
\mainmatter
\def\ECCVSubNumber{3390}  

\title{SegPGD: An Effective and Efficient Adversarial Attack for Evaluating and Boosting Segmentation Robustness} 

\titlerunning{SegPGD: An Effective and Efficient Adversarial Attack}
%
\author{Jindong Gu\inst{1,3} \and
Hengshuang Zhao\inst{2,3} \and
Volker Tresp\inst{1}\and
Philip Torr\inst{3}}
\authorrunning{J. Gu et al.}
%
\institute{$^1 \,$University of Munich \hspace{0.5cm} $^2 \,$The University of Hong Kong \\ $^3 \,$Torr Vision Group, University of Oxford \\
}
\maketitle

\begin{abstract}
Deep neural network-based image classifications are vulnerable to adversarial perturbations. The image classifications can be easily fooled by adding artificial small and imperceptible perturbations to input images. As one of the most effective defense strategies, adversarial training was proposed to address the vulnerability of classification models, where the adversarial examples are created and injected into training data during training. The attack and defense of classification models have been intensively studied in past years. Semantic segmentation, as an extension of classifications, has also received great attention recently. Recent work shows a large number of attack iterations are required to create effective adversarial examples to fool segmentation models. The observation makes both robustness evaluation and adversarial training on segmentation models challenging. In this work, we propose an effective and efficient segmentation attack method, dubbed SegPGD. Besides, we provide a convergence analysis to show the proposed SegPGD can create more effective adversarial examples than PGD under the same number of attack iterations. Furthermore, we propose to apply our SegPGD as the underlying attack method for segmentation adversarial training. Since SegPGD can create more effective adversarial examples, the adversarial training with our SegPGD can boost the robustness of segmentation models. Our proposals are also verified with experiments on popular Segmentation model architectures and standard segmentation datasets.
\end{abstract}

\section{Introduction}
Due to their vulnerability to artificial small perturbations, the adversarial robustness of deep neural networks has received great attention~\cite{szegedy2013intriguing,goodfellow2014explaining}. A large amount of attack and defense strategies have been proposed for classification in past years~\cite{cai2018curriculum,shafahi2019adversarial,wang2019bilateral,wong2020fast,zhang2019you,zheng2020efficient,vivek2020single,andriushchenko2020understanding,ye2020amata,gu2021effective,sriramanan2021towards,park2021reliably}. As an extension of classification, semantic segmentation also suffers from adversarial examples~\cite{xie2017adversarial,arnab2018robustness}. Segmentation models applied in real-world safety-critical applications also face potential threats, \textit{e.g.,} in self-driving systems~\cite{nakka2020indirect,klingner2020improved,he2019non,nesti2022evaluating,bar2020vulnerability,rossolini2022real} and in medical image analysis~\cite{daza2021towards,paschali2018generalizability,full2020studying,milletari2016v}. Hence, the adversarial robustness of segmentation has also raised great attention recently~\cite{xie2017adversarial,arnab2018robustness,xu2021dynamic,xiao2018characterizing,hendrik2017universal,tran2021robustness,lee2021adversarially,kapoor2021fourier,xiao2018characterizing,cho2020dapas,shen2019advspade,yu2021towards}.

In terms of the attack methods, different from classification, the attack goal in segmentation is to fool all pixel classifications at the same time. An effective adversarial example of a segmentation model are expected to fool as many pixel classifications as possible, which requires the larger number of attack iterations~\cite{xie2017adversarial,gu2021adversarial}. The observation makes both robustness evaluation and adversarial training on segmentation models challenging. In this work, we propose an effective and efficient segmentation attack method, dubbed SegPGD. Besides, we provide a convergence analysis to show why the proposed SegPGD can create more effective adversarial examples than PGD under the same number of attack iterations.

The right evaluation of model robustness is an important step to building robust models. Evaluation with weak or inappropriate attack methods can give a false sense of robustness~\cite{athalye2018obfuscated}. Recent work~\cite{xu2021dynamic} evaluates the robustness of segmentation models under a similar setting to the one used in classification. This could be problematic given the fact that a large number of attack iterations are required to create effective adversarial examples of segmentation~\cite{xie2017adversarial}. We evaluate the adversarially trained segmentation models in previous work with a strong attack setting, namely with a large number of attack iterations. We found the robustness can be significantly reduced. Our SegPGD can reduce the mIoU score further. For example, the mIoU of adversarially trained PSPNet~\cite{zhao2017pyramid} on Cityscapes dataset~\cite{cordts2016cityscapes} can be reduced to near zero under 100 attack iterations.

As one of the most effective defense strategies, adversarial training was proposed to address the vulnerability of classification models, where the adversarial examples are created and injected into training data during training~\cite{goodfellow2014explaining,madry2018towards}. One promising way to boost segmentation robustness is to apply adversarial training to segmentation models. However, the creation of effective segmentation adversarial examples during training can be time-consuming. In this work, we demonstrate that our effective and efficient SegPGD can mitigate this challenge. Since it can create effective adversarial examples, the application of SegPGD as the underlying attack method of adversarial training can effectively boost the robustness of segmentation models. It is worth noting that many adversarial training strategies with single-step attacks have been proposed to address the efficiency of adversarial training in classification~\cite{shafahi2019adversarial,wong2020fast,zheng2020efficient,vivek2020single,andriushchenko2020understanding}. However, they do not work well on segmentation models since the adversarial examples created by single-step attacks are not effective enough to fool segmentation models.

\noindent The contributions of our work can be summarised as follows:
\begin{itemize}
  \item Based on the difference between classification and segmentation, we propose an effective and efficient segmentation attack method, dubbed SegPGD. Especially, we show its generalization to single-step attack SegFGSM.
  \item We provide a convergence analysis to show the proposed SegPGD can create more effective adversarial examples than PGD under the same number of attack iterations.
  \item We apply SegPGD as the underlying attack method for segmentation adversarial training. The adversarial training with our SegPGD achieves state-of-the-art performance on the benchmark.
  \item We conduct experiments with popular segmentation model structures (\textit{i.e.,} PSPNet and DeepLabV3) on standard segmentation datasets (\textit{i.e.,} PASCAL VOC and Cityscapes) to demonstrate the effectiveness of our proposals.
\end{itemize}

\section{Related Work}
\noindent\textbf{Adversarial Robustness of Segmentation Models.}
The work~\cite{arnab2018robustness} makes an extensive study on the adversarial robustness of segmentation models and demonstrates the inherent robustness of standard segmentation models. Especially, they find that adversarial examples in segmentation do not transfer well across different scales and transformations. Another work~\cite{xie2017adversarial} also found that the adversarial examples created by their attack method do not transfer well across different network structures. The observations in the two works~\cite{arnab2018robustness,xie2017adversarial} indicate the standard segmentation models are inherently robust to transfer-based black-box method. The belief is broken by the work~\cite{gu2021adversarial} where they propose a method to improve the transferability of adversarial examples and show the feasibility of transfer-based black-box method. In addition, the adversarial robustness of segmentation models has also been studied from other perspectives, such as universal adversarial perturbation~\cite{hendrik2017universal,kang2020adversarial}, adversarial example detection~\cite{xiao2018characterizing}, and backdoor attack~\cite{li2021hidden}. Theses works also imply the necessity of building robust segmentation models to defend against potential threats. Along this direction, the work~\cite{klingner2020improved} shows self-supervised learning with more data can improve the robustness of standard models. However, the obtained model can be easily completely fooled with a strong attack~\cite{klingner2020improved}. A recent work~\cite{xu2021dynamic} makes the first exploration to apply adversarial training to segmentation models. We find that the adversarially trained models is still vulnerable under strong attacks. The robust accuracy of their adversarial trained models can be significantly reduced under PGD with a large number of attack iterations. In this work, we propose an effective and efficient segmentation attack method, which be used in adversarial training to build robust segmentation models against strong attacks.

\vspace{0.1cm}
\noindent\textbf{Adversarial Training of Classification Models.}
Adversarial training has been intensively studied on classification models~\cite{goodfellow2014explaining,madry2018towards}. When a multi-step attack is applied to create adversarial examples for adversarial training, the obtained model is indeed robust against various attack to some extent, as shown in~\cite{madry2018towards,wang2019bilateral,cai2018curriculum,ye2020amata}. However, adversarial training with multi-step attack can be very time consuming due to the adversarial example creation, which is N times longer than standard natural training~\cite{madry2018towards,shafahi2019adversarial}. To accelerate the adversarial training, single-step attack has also been explored therein. When standard single-step attack is applied during training, the obtained model is only robust to single-step attack~\cite{tramer2017ensemble}. One reason behind is that the gradient masking phenomenon of the model can be observed on the adversarial examples created by single-step attack. Besides, another challenge to apply single-step attack in adversarial training is the label leaking problem where the model show higher robust accuracy against single-step attack than clean accuracy~\cite{kurakin2016adversarial}. The low defensive effectiveness of single-step attack and the low efficiency of multi-step attack pose a dilemma.

One way to address the dilemma is to overcome the challenges using advanced single-step attacks~\cite{tramer2017ensemble,vivek2018gray,wang2019bilateral,zhang2019defense,wong2020fast,jia2022adversarial,jia2022boosting}, which can address label leaking problem and avoid gradient masking phenomenon. Though it boosts the robustness of the classification models, however, single-step attack based adversarial training does work well on segmentation model due to the challenge to create effective segmentation adversarial examples with a single-step attack. Another way to address the dilemma is to simulate the robustness performance of multi-step attack-based adversarial training in an efficient way~\cite{shafahi2019adversarial,zheng2020efficient,cai2018curriculum}. However, it is not clear how well the generalization of the methods above to segmentation is.

\section{SegPGD for Evaluating and Boosting Segmentation}
In semantic segmentation, given the segmentation model $f_{seg}(\cdot)$, the clean image $\boldsymbol{X}^{clean}\in\mathbb{R}^{{H}\times{W}\times{C}}$ and its segmentation label $\boldsymbol{Y}\in\mathbb{R}^{{H}\times{W}\times{M}}$, the segmentation model classifies all individual pixels of the input image $f_{seg}(\boldsymbol{X}^{clean})\in\mathbb{R}^{{H}\times{W}\times{M}}$. The notation $(H, W)$ corresponds to the size of input image, $C$ is the number of image channels, and $M$ stands for the number of output classes. The goal of the attack is to create an adversarial example to mislead classifications of all pixels of an input image.

\subsection{SegPGD: An Effective and Efficient Segmentation Attack}
Formally, the goal of attack is defined to create the adversarial example $\boldsymbol{X}^{adv}$ to mislead all the pixel classifications of an image $\boldsymbol{X}^{clean}$, \textit{i.e.,} $argmax(f_{seg}(\boldsymbol{X}^{adv})_i)\neq argmax(\boldsymbol{Y}_i)$ where $i\in [1, {H}\times{W}]$ corresponds to the index of a input pixel. One of the most popular attack method PGD~\cite{madry2018towards} creates adversarial examples via multiple iterations in Equation~\ref{equ:pgd}.

{\footnotesize
\begin{equation}
\boldsymbol{X}^{adv_{t+1}} = \phi^{\epsilon}(\boldsymbol{X}^{adv_{t}} + \alpha * \textit{sign}(\nabla_{\boldsymbol{X}^{adv_{t}}} L(f(\boldsymbol{X}^{adv_{t}}), \boldsymbol{Y}))),
\label{equ:pgd}
\end{equation}
}where $\alpha, \epsilon$ are the step size and the perturbation range, respectively. $\boldsymbol{X}^{adv_t}$ is the adversarial example after the $t$-th attack step, and the initial value is set to $\boldsymbol{X}^{adv_0} = \boldsymbol{X}^{clean}+ \mathcal{U}(-\epsilon, +\epsilon)$, which corresponds to the random initialization of perturbations. The $\phi^{\epsilon}(\cdot)$ function clips its output into the range $[\boldsymbol{X}^{clean}-\epsilon, \boldsymbol{X}^{clean}+\epsilon]$. Besides, $\boldsymbol{X}^{adv_t}$ is always clipped into a valid image space. $sign(\cdot)$ is the sign function and $\nabla_a(b)$ is the matrix derivative of b with respect to a. $L(\cdot)$ stands for the cross-entropy loss function. In segmentation, the loss is
{\footnotesize
\begin{equation}
L(f_{seg}(\boldsymbol{X}^{adv_{t}}), \boldsymbol{Y}) = \frac{1}{{H}\times{W}} \sum_{i=1}^{{H}\times{W}} CE(f_{seg}(\boldsymbol{X}^{adv_{t}})_i, \boldsymbol{Y}_i) = \frac{1}{{H}\times{W}} \sum_{i=1}^{{H}\times{W}} L_i.
\label{equ:seg_loss}
\end{equation}
}

We reformulate the loss function into two parts in Equation~\ref{equ:twoterm_loss}. The first term therein is the loss of the correctly classified pixels, while the second one is formed by the wrongly classified pixels. 
{\footnotesize
\begin{equation}
L(f_{seg}(\boldsymbol{X}^{adv_{t}}), \boldsymbol{Y}) = \frac{1}{{H}\times{W}} \sum_{j\in P^T} L_j + \frac{1}{{H}\times{W}} \sum_{k\in P^F} L_k,
\label{equ:twoterm_loss}
\end{equation}
}where $P^T$ is the set of correctly classified pixels, $P^F$ corresponds to wrongly classified ones. The two sets make up all pixels, \textit{i.e.,} $\#P^T + \#P^F = {H}\times{W}$.

The loss of the second term is often large since the wrongly classified pixels lead to large cross-entropy loss. When creating adversarial examples, the gradient of the second loss term can dominate. However, the increase of the second-term loss does not lead to better adversarial effect since the involved pixels have already been wrongly classified. To achieve highly effective adversarial examples on segmentation, a large number of attack iterations are required so that the update towards increasing the first-term loss can be accumulated to mislead correctly classified pixels.

To tackle the issue above, considering the dense pixel classifications in segmentation, we propose the \textbf{Seg}mentation-specific \textbf{PGD}, dubbed \textbf{SegPGD}, which can create more effective adversarial examples with the same number of attack iterations in Equation~\ref{equ:seg_pgd}.
{\footnotesize
\begin{equation}
L(f_{seg}(\boldsymbol{X}^{adv_{t}}), \boldsymbol{Y}) = \frac{1-\lambda}{{H}\times{W}} \sum_{j\in P^T} L_j + \frac{\lambda}{{H}\times{W}} \sum_{k\in P^F} L_k,
\label{equ:seg_pgd}
\end{equation}
}where two loss terms are weighted with $1-\lambda$ and $\lambda$, respectively. Note that the selection of $\lambda$ is non-trivial. It does not work well by simply setting $\lambda=0$ where only correctly classified pixels are considered. In such a case, the previous wrongly classified pixels can become benign again after a few attack iterations since they are ignored when updating perturbations. The claim is also consistent with the previous observation~\cite{wu2021attacking,wang2021fighting} that adversarial perturbation is also sensitive to small noise. Furthermore, setting $\lambda$ to a fixed value in [0, 0.5] does not always lead to better attack performance due to a similar reason. When most of pixel classifications are fooled after a few attack iterations, less weight on the wrongly classified pixels can make some of them benign again.

\begin{algorithm}[t]
\caption{SegPGD: An Efficient and Effective Segmentation Attack}\label{alg:seg_pgd}
\begin{algorithmic}
\footnotesize
\Require segmentation model $f_{seg}(\cdot)$, clean samples $\boldsymbol{X}^{clean}$, perturbation range $\epsilon$, step size $\alpha$, attack iterations $T$ 
\vspace{0.1cm}
\State $\boldsymbol{X}^{adv_0} = \boldsymbol{X}^{clean}+ \mathcal{U}(-\epsilon, +\epsilon)$ \Comment{initialize adversarial example}
\For{t ← 1 to T} \Comment{loop over attack iterations}
    \vspace{0.15cm}
    \State $P = f_{seg}(\boldsymbol{X}^{adv_{t-1}})$ \Comment{make predictions}
    \vspace{0.0cm}
    \State $ P^T, P^F \gets P$ \Comment{split predictions}
    \vspace{0.15cm}
    \State $\lambda(t) \gets (t - 1)/2T$ \Comment{compute weight}
    \vspace{0.05cm}
    \State $L \gets (1-\lambda(t)) * L(P^T, \boldsymbol{Y}) + \lambda(t) * L(P^F, \boldsymbol{Y})$ \Comment{loss for example updates}
    \vspace{0.15cm}
    \State $\boldsymbol{X}^{adv_{t}} \gets \boldsymbol{X}^{adv_{t-1}} + \alpha * \textit{sign}(\nabla_{\boldsymbol{X}^{adv_{t-1}}} L)$ \Comment{update adversarial examples}
    \State $\boldsymbol{X}^{adv_{t}} \gets \phi^{\epsilon}(\boldsymbol{X}^{adv_{t}})$ \Comment{clip into $\epsilon$-ball of clean image}
\EndFor
\end{algorithmic}
\end{algorithm}

In this work, instead of manually specifying a fixed value to $\lambda$, we propose to set $\lambda$ dynamically with the number of attack iterations. The intuition behind the dynamic schedule is that we mainly focus on fooling correct pixel classifications in the first a few attack iterations and then treat the wrong pixel classifications quasi equally in the last few iterations. By doing this, our SegPGD can achieve similar attack effectiveness with less iterations. We list some instances of our dynamic schedule as follows
{\footnotesize
\begin{equation}
\lambda(t) = \frac{t - 1}{2T}, \quad\quad \lambda(t) = \frac{1}{2}*\log_2 (1 + \frac{t - 1}{T}), \quad\quad \lambda(t) = \frac{1}{2}* (2^{(t - 1)/T} -1),
\label{equ:lambda}
\end{equation}
}where $t$ is the index of current attack iteration and $T$ are the number of all attack iterations. Our experiments show that all the proposed instances are similarly effective. In this work, we mainly use the first simple linear schedule. The pseudo code of our SegPGD with the proposed schedule is shown in Algorithm~\ref{alg:seg_pgd}. Further discussion on the schedules to dynamically set $\lambda$ are in Sec.~\ref{sec:exp_seg_pgd}.

Similarly, the loss function in Equation~\ref{equ:seg_pgd} can also be applied in single-step adversarial attack, \textit{e.g.,} FGSM~\cite{goodfellow2014explaining}. In the resulted SegFGSM, only correctly classified pixels are considered in case of the proposed $\lambda$ schedule. Since it only takes one-step update, the wrongly classified pixels is less likely to become benign. Hence, SegFGSM with the proposed $\lambda$ schedule (\textit{i.e.,} $\lambda=1$) also shows superior attack performance than FGSM.

In this subsection, we propose a fast segmentation attack method, \textit{i.e.,} SegPGD. It can be applied to evaluate the adversarial robustness of segmentation models in an efficient way. Besides, SegPGD can also be applied to accelerate the adversarial training on segmentation models.

\subsection{Convergence Analysis of SegPGD}
\textbf{Problem Formulation.}  The goal of the attack is to create an adversarial example $\boldsymbol{X}^{adv}$ to maximize cross-entropy loss of all the pixel classifications. The adversarial example is constrained into $\epsilon$-ball of the clean example $\boldsymbol{X}^{clean}$. The cross-entropy loss of $i$-th pixel is 
{\footnotesize
\begin{equation} 
L(\boldsymbol{X}, \boldsymbol{Y}_i) =  CE(f_{seg}(\boldsymbol{X}^{adv_{t}})_i, \boldsymbol{Y}_i).
\end{equation}
}

The process to create adversarial example for segmentation can be formulated into a constrained minimization problem
{\footnotesize
\begin{equation}
\min_{\boldsymbol{X}} \frac{1}{{H}\times{W}} \sum_{i=1}^{H\times W} {g_i(\boldsymbol{X})} \quad s.t. \; \| \boldsymbol{X}-\boldsymbol{X}^{clean} \|_{\infty} < \epsilon \; \mathrm{and} \; \boldsymbol{X} \in [0, 1],
\end{equation}
}where $g_i(\boldsymbol{X}) = - L(\boldsymbol{X}, \boldsymbol{Y}_i).$ The variable is constrained into concave region since both constraints are linear.

Projected Gradient Descent-based optimization method is often applied to solve the constrained minimization problem above~\cite{madry2018towards}. The method first takes a step towards the negative gradient
direction to get a new point while ignoring the constraint, and then correct the new point by projecting it back into the constraint set. 

The gradient-descent step of PGD attack is
\begin{equation}
\boldsymbol{X}^{t+1} = \boldsymbol{X}^t - \alpha * \textit{sign}(\nabla \sum_{i=1}^{H\times W} g_i(\boldsymbol{X}^t)),
\end{equation} In contrast, the gradient-descent step of our SegPGD attack is
\begin{equation}
\boldsymbol{X}^{t+1} = \boldsymbol{X}^t - \alpha * \textit{sign}(\nabla (\sum_{j\in P^T} (1-\lambda(t)) g_j(\boldsymbol{X}^t) + \sum_{k\in P^F} \lambda(t) g_k(\boldsymbol{X}^t))),
\end{equation}where $\alpha$ is the step size. The initial point is the original clean example $\boldsymbol{X}^{clean}$ or a random initialization $\boldsymbol{X}^{clean}+ \mathcal{U}(-\epsilon, +\epsilon)$.   \vspace{0.1cm}

\begin{figure}[t]
    \begin{subfigure}[b]{0.48\textwidth}
    \centering
    \includegraphics[scale=0.53]{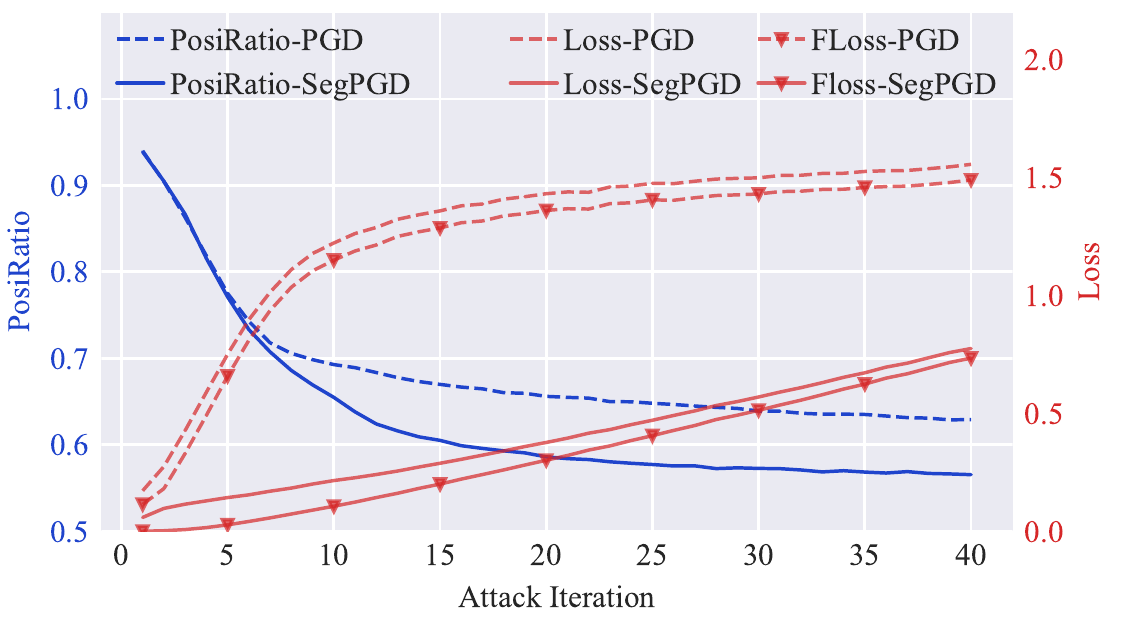}\vspace{-0.1cm} 
    \caption{\scriptsize PGD3 AT-PSPNet on VOC}
    \label{fig:conv_ana_at}
    \end{subfigure} \hspace{0.05cm} 
    \begin{subfigure}[b]{0.48\textwidth}
    \centering
    \includegraphics[scale=0.53]{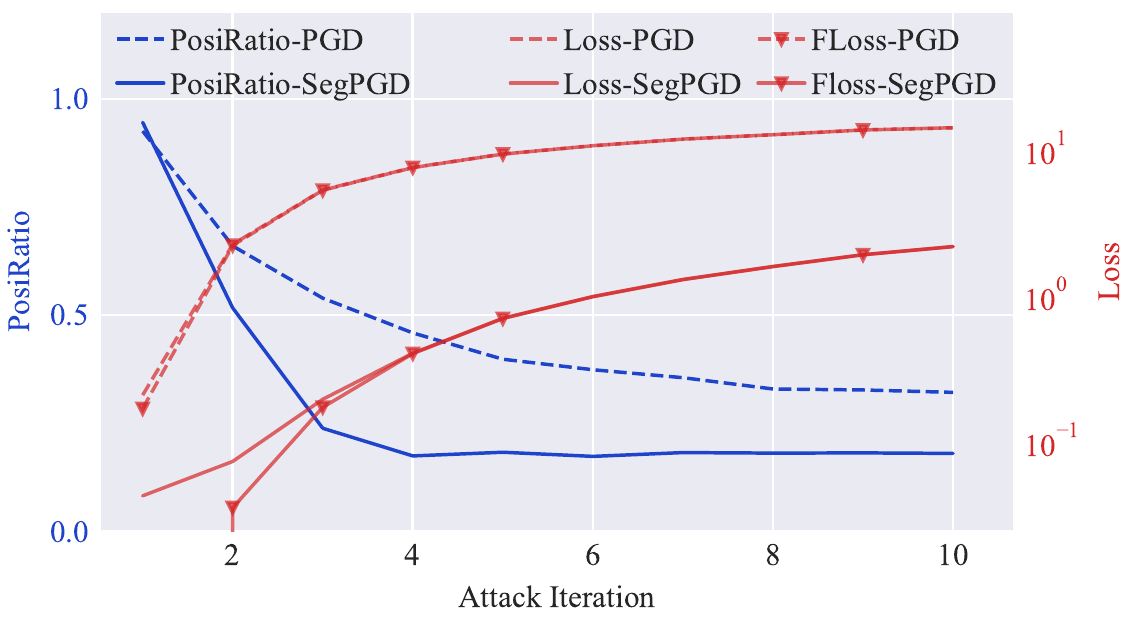}\vspace{-0.1cm} 
    \caption{\scriptsize Standard PSPNet on VOC}
    \label{fig:conv_ana_std}
    \end{subfigure}
    \caption{Convergence Analysis. SegPGD marked with blue solid lines achieve higher MisRatio than PGD under the same number of attack iterations. The loss of false classified pixels (\textbf{FLoss}) marked with triangle down dominate the overall loss (i.e. red lines without markers) during attacks. Compared to PGD, the FLoss in SegPDG makes up a smaller portion of the overall loss since SegPGD main focuses on correctly classified pixels in the first a few attack iterations.}
    \label{fig:conv_ana}
\end{figure}

\noindent\textbf{Convergence Criterion.} In classification task, the loss is directly correlated with attack goal. The larger the loss is, the more likely the input is to be misclassified. However, it does not hold in segmentation task. The large loss of segmentation not necessarily leads to more pixel misclassifications since the loss consists of losses of all pixel classifications. Once a pixel is misclassified, the increase of the loss on the pixel does not bring more adversarial effect. Hence, we propose a new convergence  criterion for segmentation, dubbed MisRatio, which is defined as the ratio of misclassified pixels to all input pixels. \vspace{0.1cm}

\noindent\textbf{Convergence Analysis.} In the first step to update adversarial examples, the update rule of our SegPGD can be simplified as 
\begin{equation}
\boldsymbol{X}^1 = \boldsymbol{X}^0 + \alpha * \textit{sign}(\sum_{j\in P^T} \nabla g_j(\boldsymbol{X}^t)),
\end{equation}For almost all misclassified pixels $k\in P^F$ of $\boldsymbol{X}^0$, the $k$-th pixel of $\boldsymbol{X}^1$ is still misclassified since natural misclassifications are not sensitive to small adversarial noise in general. The claim is also true with PGD update rule. Besides, our SegPGD can turn part of the pixels $k\in P^T$ of $\boldsymbol{X}^0$ into misclassified ones of $\boldsymbol{X}^1$. However, PGD is less effective to do so since the update direction also takes the misclassified pixels of $\boldsymbol{X}^0$ into consideration. Therefore, our SegPGD can achieve higher MisRatio than PGD in the first step.

In all intermediate steps, both SegPGD and PGD leverage gradients of all pixels classification loss to update adversarial examples. The difference is that our SegPGD assign more weight to loss of correctly classified pixel classifications. The assigned value depends on the update iteration $t$. Our SegPGD focuses more on fooling correctly classified pixels at first a few iterations and then treat both quasi equally. By doing this, our SegPGD can achieve higher MisRatio than PGD under the same attack iterations.

In Fig.~\ref{fig:conv_ana}, we show the pixel classification loss and PosiRatio (=1 - MisRatio) in each attack iteration. Fig.~\ref{fig:conv_ana_at} shows the case to attack adversarially trained PSPNet on VOC (see more details in experimental section). SegPGD marked with blue solid lines achieve higher MissRatio than PGD under the same number of attack iterations. The loss of False classified pixels (FLoss) marked with triangle down dominate the overall loss (i.e. red lines without markers) during attacks. Compared to PGD, the FLoss in SegPDG makes up a smaller portion of the overall loss since SegPGD main focuses on correctly classified pixels in the first a few attack iterations. Note that the scale of loss does not matter since only the signs of input gradients are leveraged to create adversarial examples.

\begin{algorithm}[t]
\caption{Segmentation Adversarial Training with SegPGD}\label{alg:hcat}
\begin{algorithmic}
\footnotesize
\Require segmentation model $f_{seg}(\cdot)$, training iterations $\mathcal{N}$, perturbation range $\epsilon$, step size $\alpha$, attack iterations $T$ 
\vspace{0.1cm}
\For{i ← 1 to $\mathcal{N}$}
    \vspace{0.15cm}
    \State $\boldsymbol{X}_1^{clean}$, $\boldsymbol{X}_2^{clean}$ ← $\boldsymbol{X}^{clean}$ \Comment{split mini-batch}
    \vspace{0.05cm}
    \State $\boldsymbol{X}_2^{adv}$ ← SegPGD($f_{seg}(\cdot)$, $\boldsymbol{X}_2^{clean}$, $\epsilon$, $\alpha$, i) \Comment{create adversarial examples}
    \vspace{0.05cm}
    \State $L$ ← $L$($f_{seg}(\boldsymbol{X}_1^{clean})$,$\boldsymbol{Y}_1$) + $L$($f_{seg}(\boldsymbol{X}_2^{adv})$,$\boldsymbol{Y}_2$) \Comment{loss for network updates}
\EndFor
\end{algorithmic}
\label{alg:at_segpgd}
\end{algorithm} \vspace{-0.1cm}

\subsection{Segmentation Adversarial Training with SegPGD}
Adversarial training, as one of the most effective defense methods, has been well studied in the classification task. In classification, the main challenge of applying adversarial training is computational cost. It requires multiple gradient propagation to produce adversarial images, which makes adversarial training slow. In fact, it can take 3-30 times longer to train a robust
network with adversarial training than training a non-robust equivalent~\cite{shafahi2019adversarial}. The segmentation task makes the adversarial training more challenging. More attack iterations are required to create effective adversarial examples for boosting segmentation robustness. \textit{E.g.,} more than 100 attack iterations are required to fool segmentation~\cite{xie2017adversarial}.

In this work, we improve segmentation adversarial training by applying SegPGD as the underlying attack. As an effective and efficient segmentation attack method, SegPGD can create more effective adversarial examples than the popular PGD. By injecting the created adversarial examples into the training data, adversarial training with SegPGD can achieve a more robust segmentation model with the same computational cost. Following the previous work, the adversarial training procedure on segmentation is shown in Algorithm \ref{alg:at_segpgd}.

\section{Experiment}
In this section, we first introduce the experimental setting. Then, we show the effectiveness of SegPGD. Specifically, we show SegPGD can achieve similar attack effect with less attack iterations than PGD on both standard models and adversarially trained models. In the last part, we show that adversarial training with SegPGD can achieve more adversarially robust segmentation models.

\subsection{Experimental Setting}
\textbf{Datasets.} The popular semantic segmentation datasets, PASCAL VOC 2012 (VOC)~\cite{everingham2010pascal} and Cityscapes (CS)~\cite{cordts2016cityscapes}, are adopted in experiments. VOC dataset contains 20 object classes and one class for background, with 1,464, 1,499, and 1,456 images for training, validation, and testing, respectively. Following the popular protocol~\cite{hariharan2015hypercolumns}, the training set is augmented to 10,582 images. Cityscapes dataset contains urban scene understanding images with 19 categories, which contains high-quality pixel-level annotations with 2,975, 500, and 1,525 images for training, validation, and testing, respectively.

\vspace{0.1cm}
\noindent\textbf{Models.} We choose popular semantic segmentation architectures PSPNet~\cite{zhao2017pyramid} and DeepLabv3~\cite{chen2017rethinking} for our experiments. The standard configuration of the model architectures is used as in~\cite{zhao2017pyramid}. By default, ResNet50~\cite{he2016deep} is applied as a backbone for feature extraction in both segmentation models.

\vspace{0.1cm}
\noindent\textbf{Adversarial Attack.} We choose the popular single-step attack FGSM~\cite{goodfellow2014explaining} and the popular multiple-step attack PGD~\cite{madry2018towards} as our baseline attack methods. In this work, we focus on $\ell_{\infty}$-based perturbations. The maximum allowed perturbation value $\epsilon$ is set to 0.03 = 8/255. The step size $\alpha$ is set to 0.03 for FGSM and 0.01 for PGD. The PGD with 3 attack iterations is denoted as PGD3. Besides, for evaluating the robustness of segmentation models, we also apply attack methods, such as CW attack~\cite{carlini2017towards}, DeepFool~\cite{moosavi2016deepfool} and $\ell_{2}$-based BIM~\cite{kurakin2016adversarial}.

\vspace{0.1cm}
\noindent\textbf{Metrics.} The standard segmentation evaluation metric mIoU (in \%) is used to evaluate the adversarial robustness of segmentation models. The mIoUs on both clean image and adversarial images are reported, respectively. The higher the mIoUs are, the more robust the model is.

\begin{figure}[!t]
\centering
    \begin{subfigure}[b]{0.23\textwidth}
    \centering
    \includegraphics[scale=0.47]{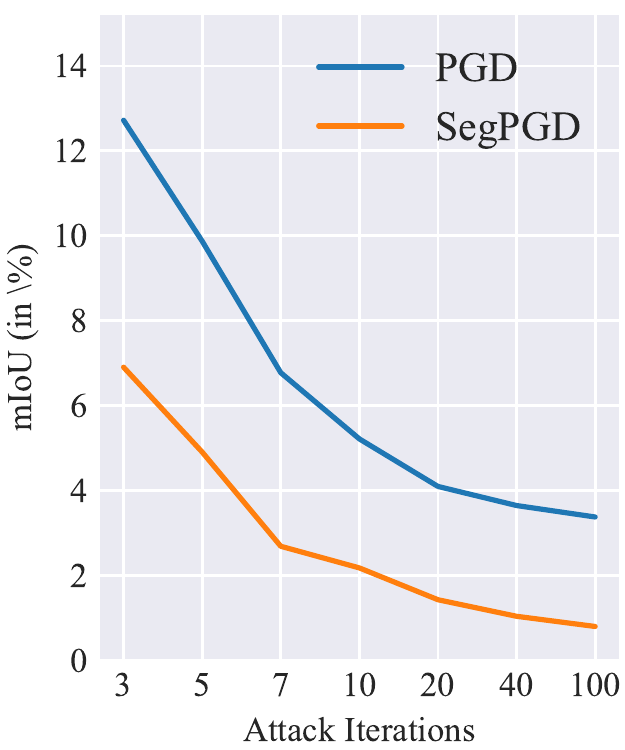}\vspace{-0.15cm}  
    \caption{\scriptsize Standard PSPNet}
    \end{subfigure} 
    \begin{subfigure}[b]{0.26\textwidth}
    \centering
    \includegraphics[scale=0.47]{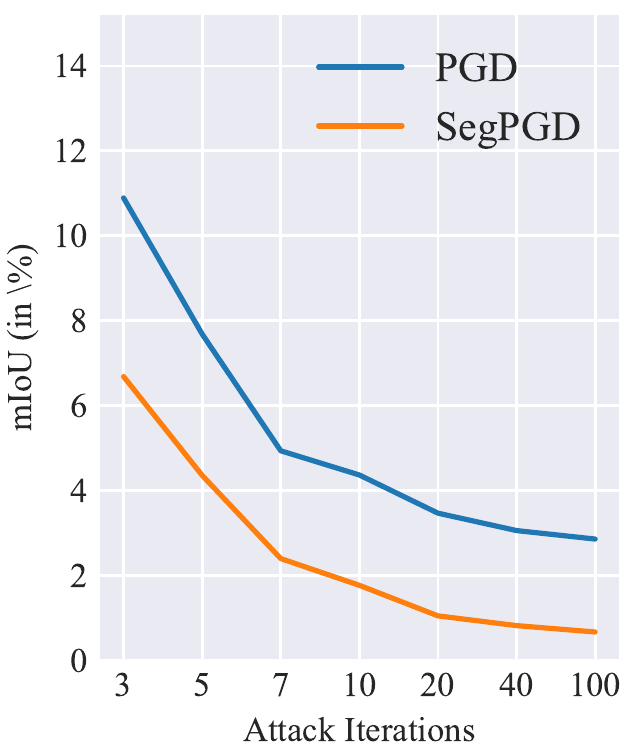}\vspace{-0.15cm} 
    \caption{\scriptsize Standard DeepLabV3}
    \end{subfigure}  \hspace{-0.2cm}
    \begin{subfigure}[b]{0.23\textwidth}
    \centering
    \includegraphics[scale=0.47]{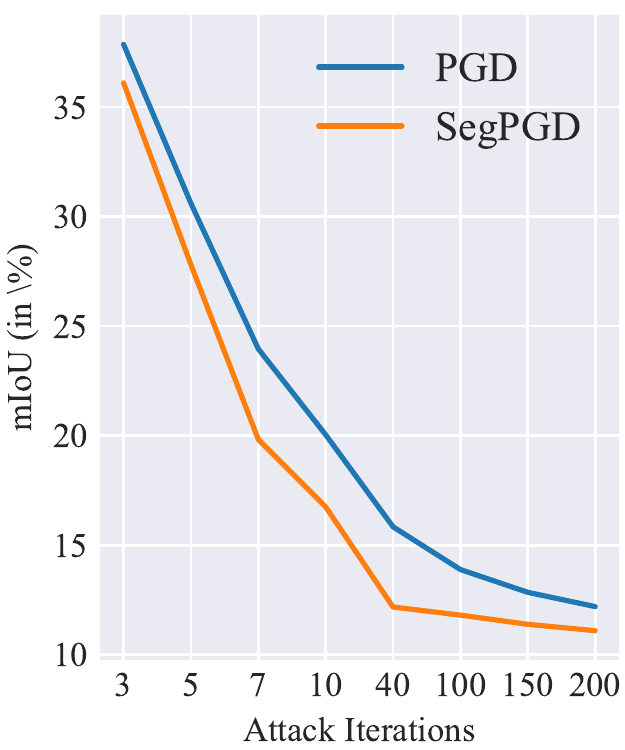}\vspace{-0.15cm} 
    \caption{\scriptsize AT PSPNet}
    \end{subfigure} \hspace{0.05cm}
    \begin{subfigure}[b]{0.23\textwidth}
    \centering
    \includegraphics[scale=0.47]{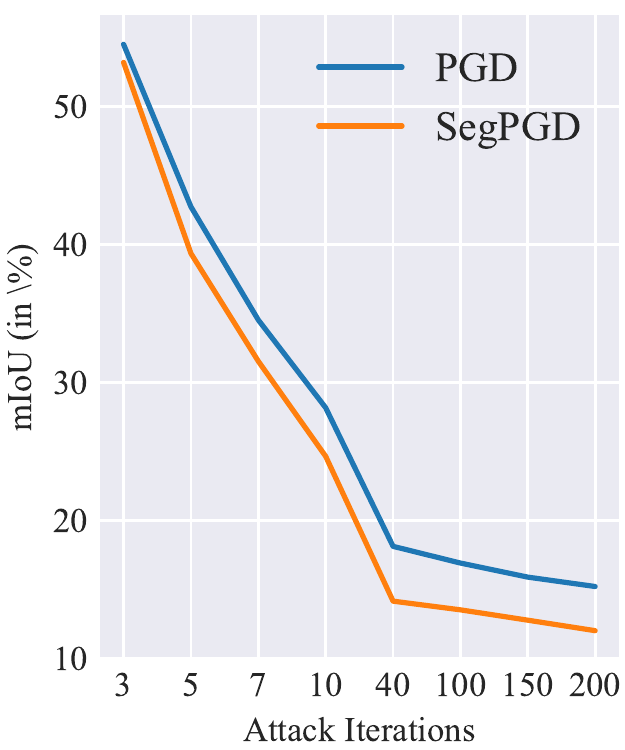}\vspace{-0.15cm} 
    \caption{\scriptsize AT DeepLabV3}
    \end{subfigure} \vspace{0.1cm} 
    
    \begin{subfigure}[b]{0.23\textwidth}
    \centering
    \includegraphics[scale=0.47]{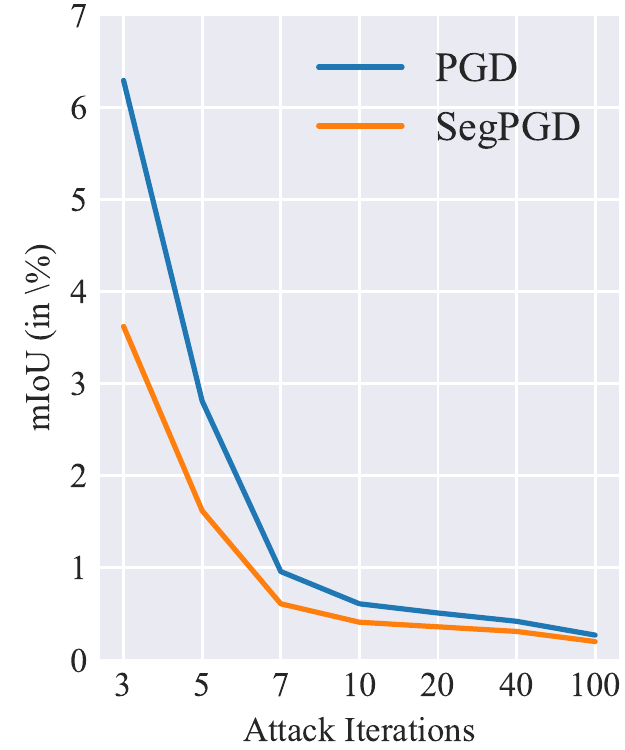}\vspace{-0.15cm}  
    \caption{\scriptsize Standard PSPNet}
    \end{subfigure} 
    \begin{subfigure}[b]{0.26\textwidth}
    \centering
    \includegraphics[scale=0.47]{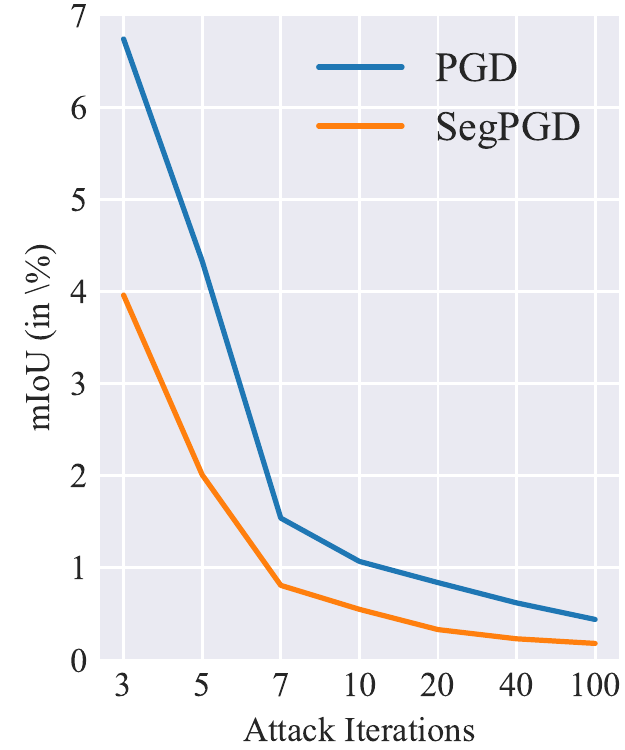}\vspace{-0.15cm} 
    \caption{\scriptsize Standard DeepLabv3}
    \end{subfigure}  \hspace{-0.2cm}
    \begin{subfigure}[b]{0.23\textwidth}
    \centering
    \includegraphics[scale=0.47]{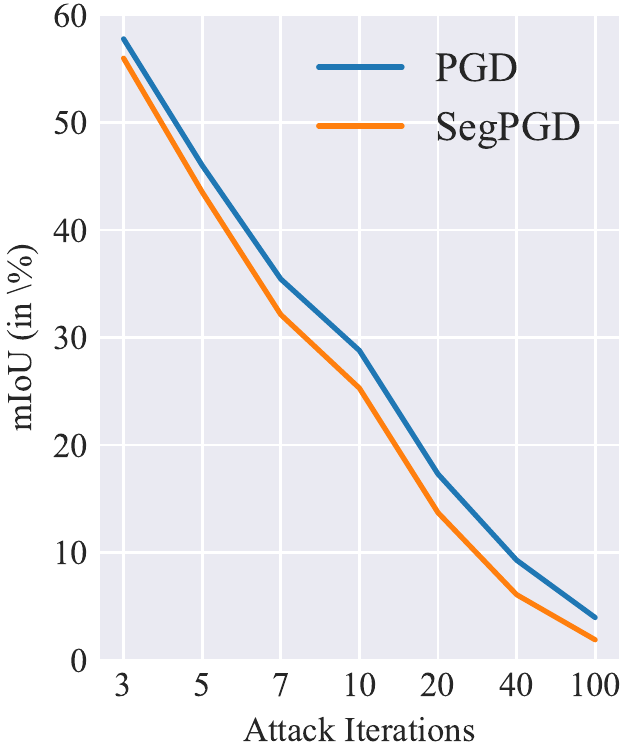}\vspace{-0.15cm} 
    \caption{\scriptsize AT PSPNet}
    \end{subfigure}  \hspace{0.05cm}
    \begin{subfigure}[b]{0.23\textwidth}
    \centering
    \includegraphics[scale=0.47]{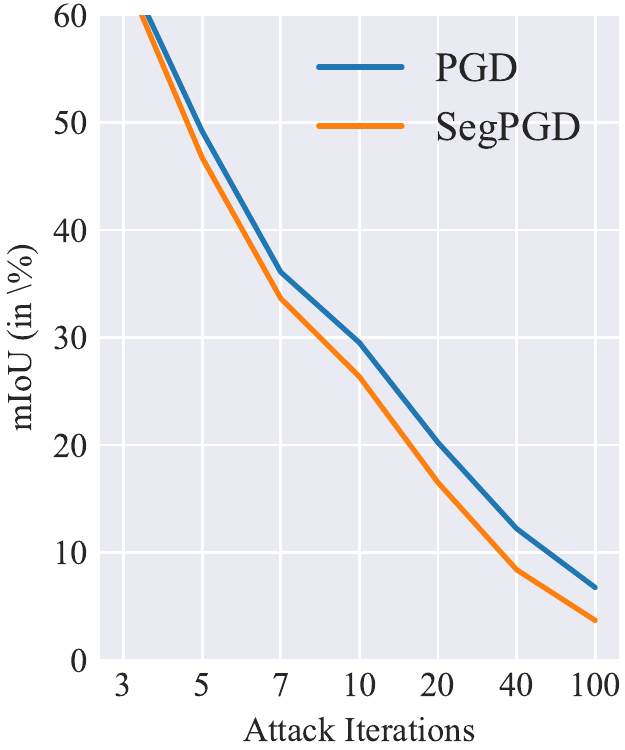}\vspace{-0.15cm}
    \caption{\scriptsize AT DeepLabV3}
    \end{subfigure} \vspace{0.1cm}

    \caption{SegPGD is more effective and efficient than PGD. SegPGD creates more effective adversarial examples with the same number of attack iterations and converges to a better minima than PGD. The subfigures (a-d) show the segmentation mIoUs on VOC, while the scores on Cityscapes are reported in the subfigures (e-h). AT PSPNet stands for the adversarially trained PSPNet.}\vspace{-0.2cm}
    \label{fig:quanti_eval}
\end{figure}

\subsection{Evaluating Segmentation Robustness with SegPGD}
\label{sec:exp_seg_pgd}
\textbf{Quantitative Evaluation.} We train PSPNet and DeepLabV3 on VOC and Cityscapes, respectively. Both standard training and adversarial training are considered in this experiment. PGD with 3 attack iterations is applied as the underlying attack method of adversarial training. This result in 8 models. We apply PGD and SegPGD on the 8 models. On each model, we report the final mIoU under attack with different attack iterations, \textit{e.g.,} 20, 40 and 100. As shown in Fig.~\ref{fig:quanti_eval}, the segmentation models show low mIoU on the adversarial examples created by our SegPGD. SegPGD achieve can converge faster to a better minima than PGD, which shows the high effectiveness and efficiency of SegPGD.

\begin{figure}[t]
    \begin{subfigure}[b]{\textwidth}
    \centering
    \includegraphics[scale=0.145]{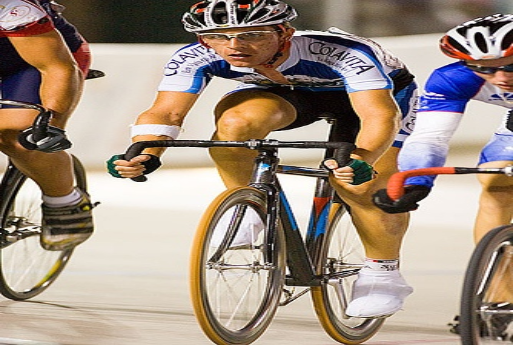} \hspace{0.03cm} \includegraphics[scale=0.145]{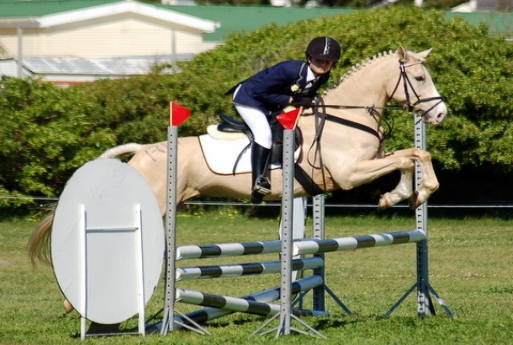} \hspace{0.03cm}
    \includegraphics[scale=0.145]{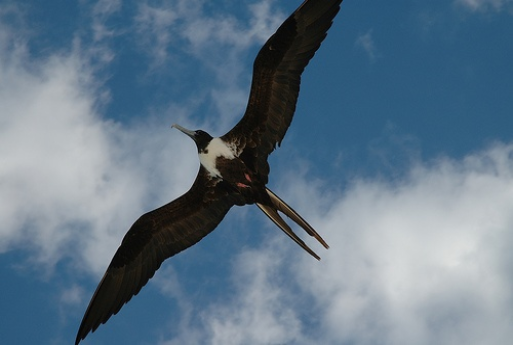} \hspace{0.03cm}
    \includegraphics[scale=0.145]{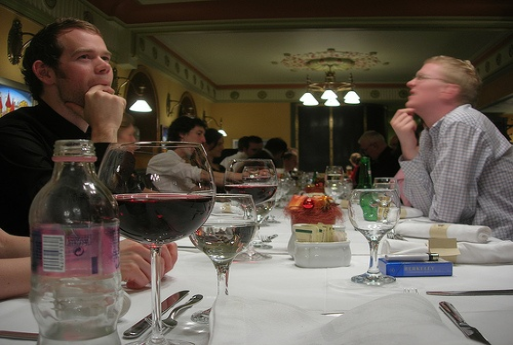} \hspace{0.03cm}
    
    \includegraphics[scale=0.145]{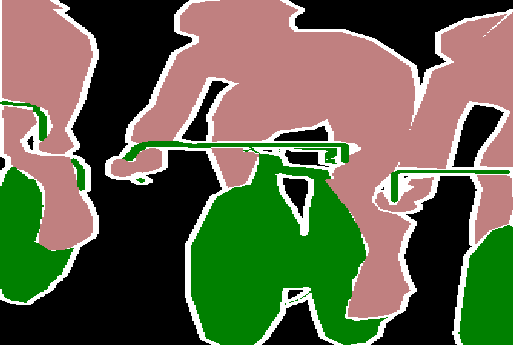} \hspace{0.03cm} \includegraphics[scale=0.145]{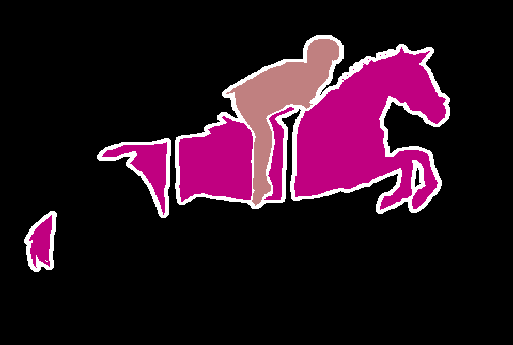} \hspace{0.03cm}
    \includegraphics[scale=0.145]{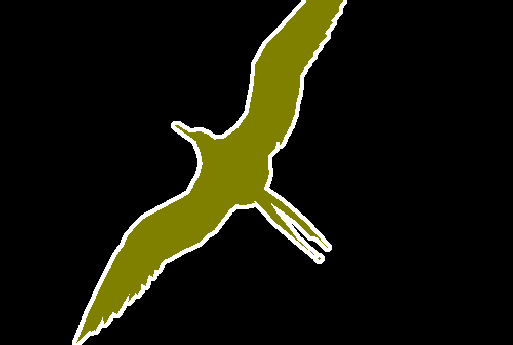} \hspace{0.03cm}
    \includegraphics[scale=0.145]{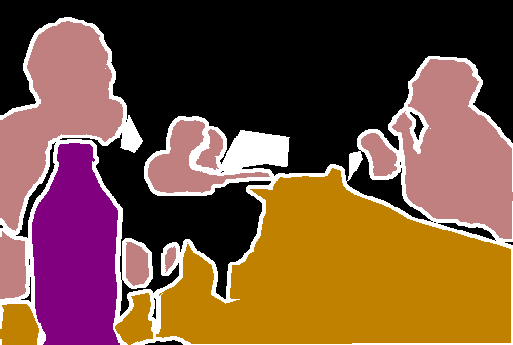} \vspace{-0.2cm}\hspace{0.03cm} 
    \caption{\scriptsize Clean Images and Ground-truth Masks} \vspace{0.2cm}
    \label{subfig:img_gt}
    \end{subfigure} 
    \begin{subfigure}[b]{\textwidth}
    \centering
    \includegraphics[scale=0.145]{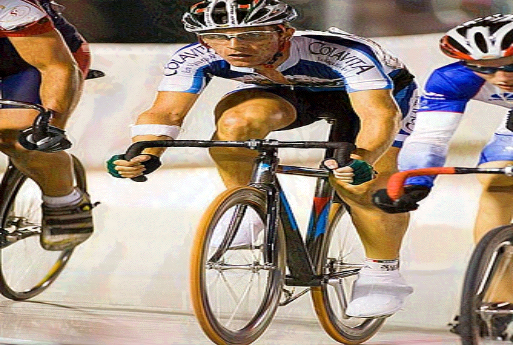} \hspace{0.03cm} \includegraphics[scale=0.145]{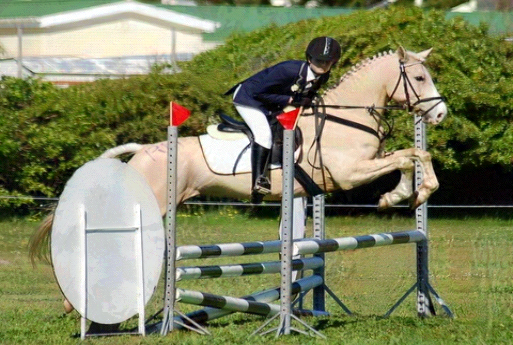} \hspace{0.03cm}
    \includegraphics[scale=0.145]{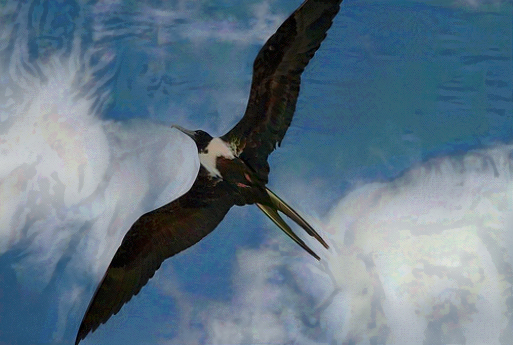} \hspace{0.03cm}
    \includegraphics[scale=0.145]{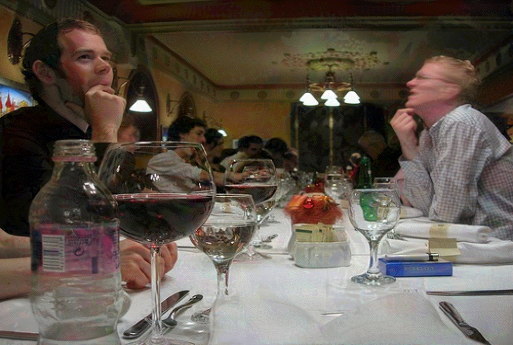} \hspace{0.03cm}
    
    \includegraphics[scale=0.145]{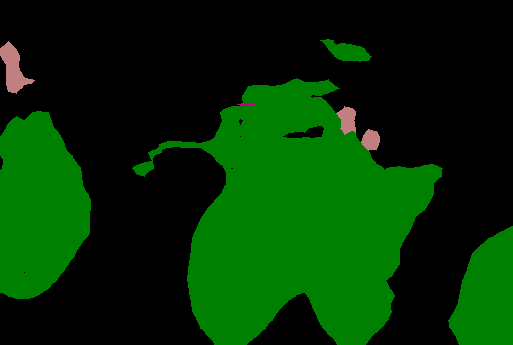} \hspace{0.03cm} \includegraphics[scale=0.145]{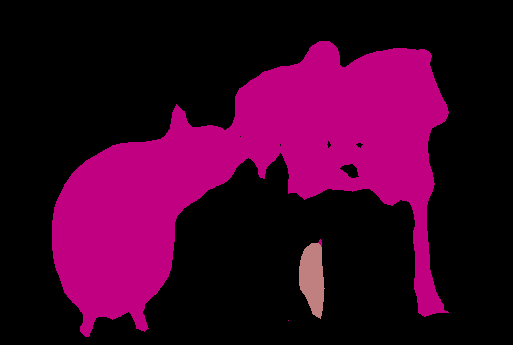} \hspace{0.03cm}
    \includegraphics[scale=0.145]{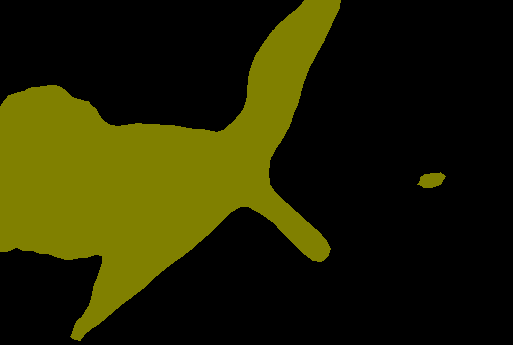} \hspace{0.03cm}
    \includegraphics[scale=0.145]{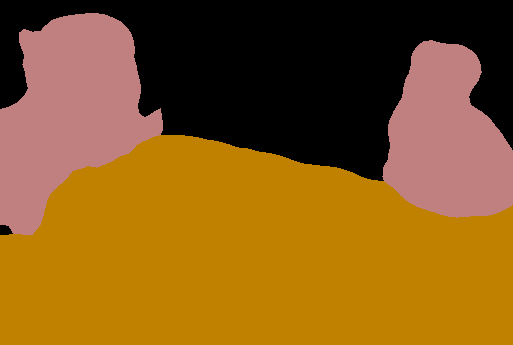} \vspace{-0.2cm}\hspace{0.03cm} 
    \caption{\scriptsize Adversarial Images Created by PGD and Model Predictions} \vspace{0.2cm}
    \label{subfig:pgd_adv}
    \end{subfigure} \vspace{0.0cm}
    \begin{subfigure}[b]{\textwidth}
    \centering
    \includegraphics[scale=0.145]{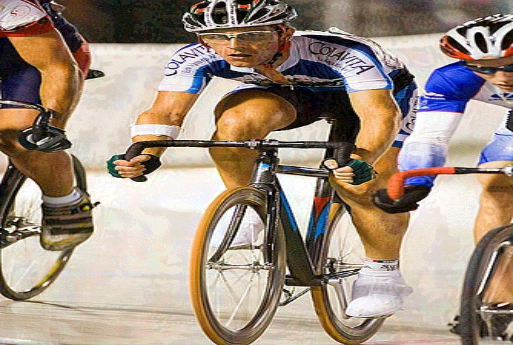} \hspace{0.03cm} \includegraphics[scale=0.145]{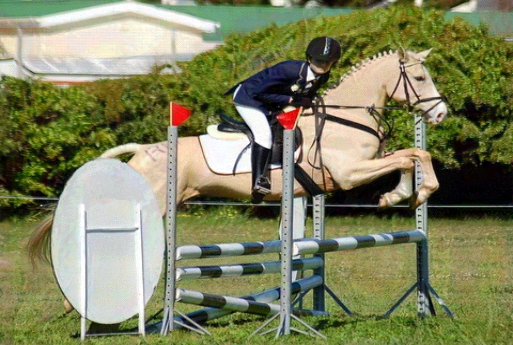} \hspace{0.03cm}
    \includegraphics[scale=0.145]{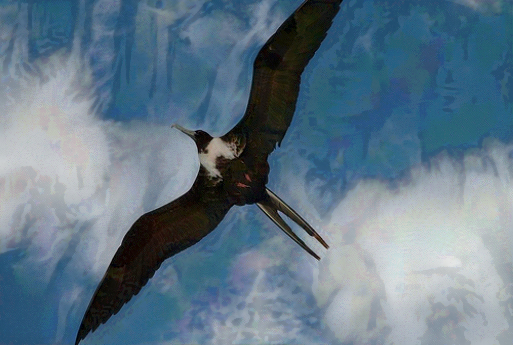} \hspace{0.03cm}
    \includegraphics[scale=0.145]{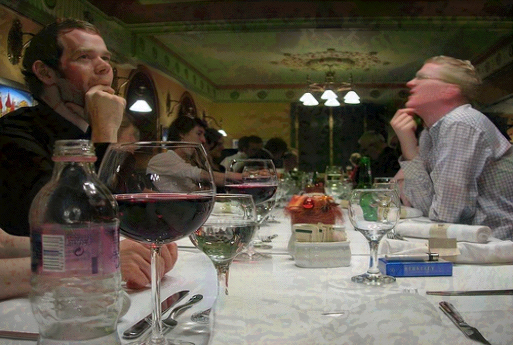} \hspace{0.03cm}
    
    \includegraphics[scale=0.145]{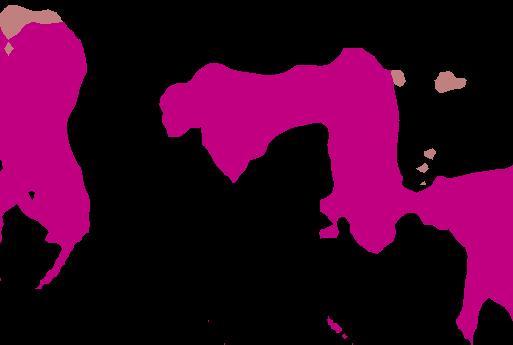} \hspace{0.03cm} \includegraphics[scale=0.145]{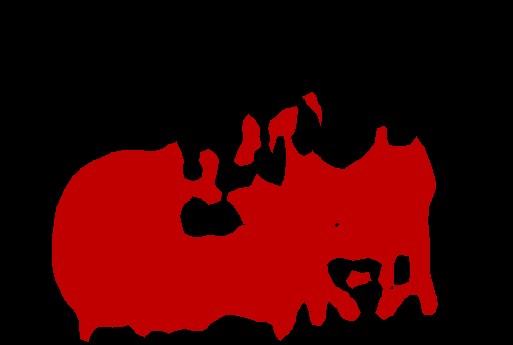} \hspace{0.03cm}
    \includegraphics[scale=0.145]{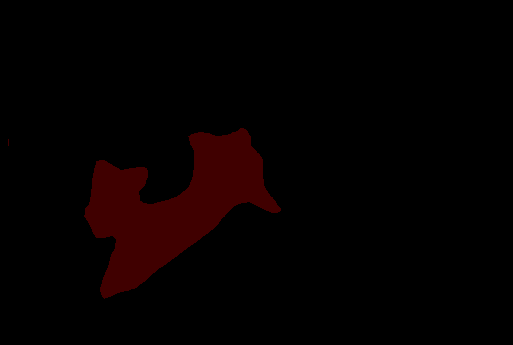} \hspace{0.03cm}
    \includegraphics[scale=0.145]{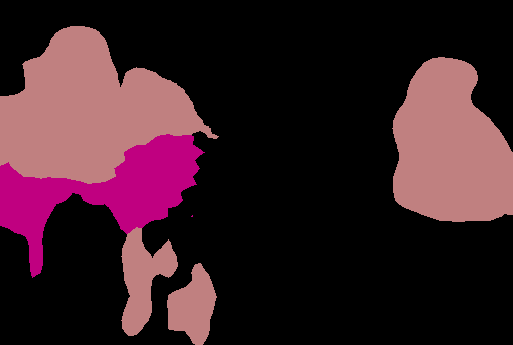} \vspace{-0.2cm}\hspace{0.03cm} 
    \caption{\scriptsize Adversarial Images Created by SegPGD and Model Predictions} \vspace{0.0cm}
    \label{subfig:segpgd_adv}
    \end{subfigure}
    \caption{Visualizing of Adversarial Examples and Predictions on them. SegPGD create more effective adversarial examples than PGD.} \vspace{-0.4cm}
    \label{fig:quali_eval}
\end{figure}

\vspace{0.1cm}
\noindent\textbf{Qualitative Evaluation.} For qualitative evaluations, we visualize the created adversarial examples and model's predictions on them. We take the adversarial examples created on standard PSPNet on VOC with 20 attack iterations as examples. As shown in Fig.~\ref{fig:quali_eval}, the adversarial perturbations created by both PGD and SegPGD are imperceptible to human vision. In other words, the created adversarial examples in Fig.~\ref{subfig:pgd_adv} and~\ref{subfig:pgd_adv} are not distinguishable from the counter-part clean images in Fig.~\ref{subfig:img_gt}. The predicted masks on the adversarial examples by SegPGD have deviated more from the ground truth than the ones corresponding to PGD. The visualization in Fig.~\ref{fig:quali_eval} shows SegPGD creates more effective adversarial examples than PGD under the same number of attack iterations.

\vspace{0.1cm}
\noindent\textbf{Comparison with other Segmentation Attack Methods.}
The segmentation attack methods have also been explored in related work. The work~\cite{hendrik2017universal} aim to create adversarial perturbations that are always deceptive when added to any sample. Similarly, The work~\cite{kang2020adversarial} creates universal perturbations to attack multiple segmentation models. Since more constraints are applied to universal perturbations, both types of universal adversarial perturbations are supposed to be less effective than the sample-specific ones. Another work~\cite{hendrik2017universal} related to us proposes Dense Adversary Generation (DAG), which can be seen as a special case of our SegPGD along with other minor differences. DAG only considers the correctly classified pixels in each attack iteration, which is equivalent to set $\lambda = 0$ in our SegPGD. To further improve the attack effectiveness, the work~\cite{gupta2019mlattack} proposes multiple-layer attack (MLAttack) where the losses in feature spaces of multiple intermediate layers and the one in the final output layer are combined to create adversarial examples. SegPGD outperforms both DAG and MLAttack in terms of both efficiency and effectiveness, as shown in Appendix A.

\vspace{0.1cm}
\noindent\textbf{Single-Step Attack.}
When a single attack iteration is applied, SegPGD is degraded to SegFGSM. In SegFGSM, only the loss of correctly classified pixels are considered in the case of the proposed $\lambda$ schedule. We compare FGSM and SegFGSM and report the mIOU. Our SegFGSM outperforms FGSM on both standard models and adversarially trained models. See Appendix B for details.

\begin{figure}[t]
    \begin{subfigure}[b]{0.46\textwidth}
    \centering
    \includegraphics[scale=0.52]{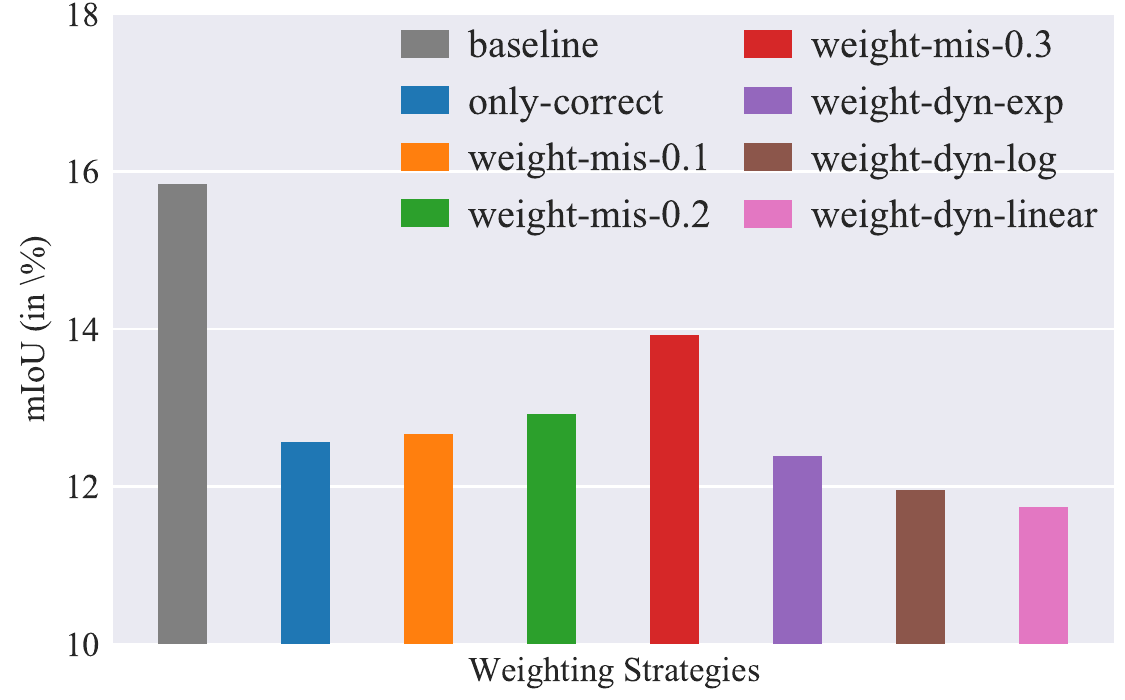}\vspace{-0.1cm} 
    \caption{\scriptsize Standard PSPNet on VOC}
    \end{subfigure} \hspace{0.1cm} 
    \begin{subfigure}[b]{0.46\textwidth}
    \centering
    \includegraphics[scale=0.52]{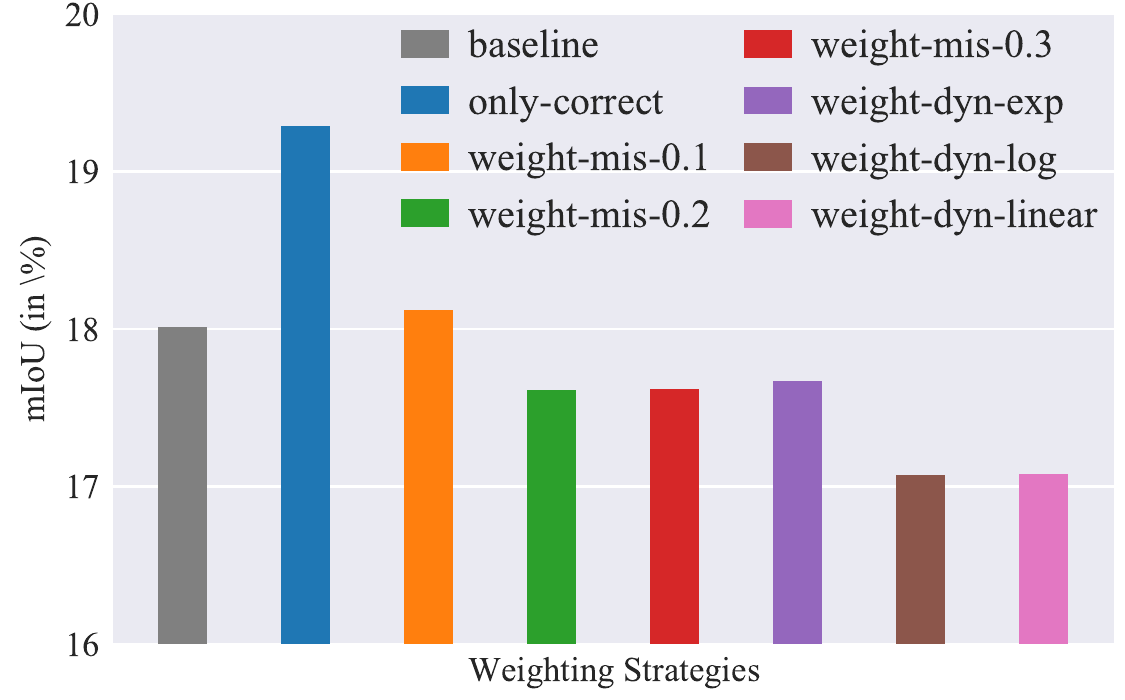}\vspace{-0.1cm} 
    \caption{\scriptsize PGD3 AT-PSPNet on VOC}
    \end{subfigure}
    \caption{Schedules for weighting misclassified pixels. SegPGD with our \textit{weight-dyn-linear} weighting schedules can better reduce the mIoU and achieve better attack effectiveness than the ones with baseline schedules.}
    \label{fig:weight_schedues}
\end{figure}

\begin{table}[t]
\begin{center}
\caption{Adversarial Training on VOC Dataset. We evaluate the robustness of adversarially trained models with various attacks, especially under strong attacks (\textit{e.g.,} PGD with 100 attack iterations). We report mIoU scores on different segmentation architectures and different adversarial training settings. Adversarial training with our SegPGD can boost the robustness of segmentation models.}
\label{tab:seg_base_voc}
\scriptsize
\setlength\tabcolsep{0.15cm}
\begin{tabular}{c | cc ccc cccc}
\toprule
PSPNet & Clean & CW & DeepFool & BIMl2  & PGD10 & PGD20 & PGD40 & PGD100  \\
\midrule
Standard & 76.64  & 4.72 & 14.2 & 15.32 & 5.21 & 4.09 & 3.64 & 3.37  \\
\midrule
DDCAT~\cite{xu2021dynamic} & 75.89  & 39.73 & \textbf{67.76} & 45.36 & 28.46 & 19.03 & 14.30 & 10.51  \\
PGD3-AT & 74.51 & 52.23 & 55.46 & \textbf{51.56} & 20.04 & 17.34 & 15.84 & 13.89 \\
\textbf{SegPGD3-AT} & 75.38 & \textbf{56.52} & 59.47 & 50.17 & \textbf{26.6} & \textbf{20.69} & \textbf{17.19} & \textbf{14.49}   \\
\midrule
PGD7-AT & 74.99 & 42.30 & 45.05 & 47.21 & 21.79 & 19.39 & 17.99 & 16.97 \\
\textbf{SegPGD7-AT} & 74.45 & \textbf{48.79} & \textbf{51.44} & 45.15 & \textbf{25.73} & \textbf{22.05} & \textbf{20.61} & \textbf{19.23} \\
\midrule
\toprule
DeepLabv3 & Clean & CW & DeepFool & BIMl2 & PGD10 & PGD20 & PGD40 & PGD100 \\
\midrule
Standard & 77.36 & 5.24 & 13.57 & 14.76 & 4.36 & 3.46 & 3.05 & 2.85 \\
\midrule
DDCAT~\cite{xu2021dynamic} & 74.76  & \textbf{66.09} & \textbf{67.20} & 37.18 & 23.36 & 15.93 & 11.65 & 8.62  \\
PGD3-AT & 75.03 & 57.10 & 60.23 & 36.83 & \textbf{28.16} & 20.77 & 18.12 & 16.91  \\
\textbf{SegPGD3-AT} & 75.01 & 59.55 & 62.12 & \textbf{39.46} & 26.29 & \textbf{20.92} & \textbf{19.1} & \textbf{18.24} \\
\midrule
PGD7-AT & 73.45 & 48.51 & 48.87 & 43.13 & 26.23 & 21.15 & 20.06 & 19.10 \\
\textbf{SegPGD7-AT} & 74.46 & \textbf{51.42} & \textbf{51.47} & 42.91 & \textbf{30.95} & \textbf{26.68} & \textbf{24.32} & \textbf{23.09} \\
\bottomrule
\end{tabular} \vspace{-0.4cm}
\end{center}
\end{table}

\begin{table}[t]
\begin{center}
\caption{Adversarial Training on Cityscapes Dataset. This table show that the boosting effect of adversarial training with our SegPGD still clearly holds on a different dataset. Besides, we show the previous adversarially trained baseline model can be reduced to near zero under strong attack, \textit{i.e.,} PGD3-AT PSPNet under PGD100 attack. Our segPGD improves the robustness significantly.}
\label{tab:seg_base_city}
\scriptsize
\setlength\tabcolsep{0.15cm}
\begin{tabular}{c | c  c ccc cccc}
\toprule
PSPNet & Clean & CW & DeepFool & BIMl2 & PGD10 & PGD20 & PGD40 & PGD100 \\
\midrule
Standard & 73.98 & 5.94 & 12.68 & 12.36 & 0.96 & 0.61 & 0.42 & 0.27 \\
\midrule
DDCAT~\cite{xu2021dynamic} & 71.86  & \textbf{53.19} & \textbf{54.08} & \textbf{48.86} & 24.40 & 20.90 & 17.93 & 12.97  \\
PGD3-AT & 71.28 & 35.21 & 36.84 & 32.22 & 28.79 & 17.3 & 9.29 & 3.95   \\
\textbf{SegPGD3-AT} & 71.01 & 36.30 & 38.27 & 35.34 & \textbf{33.52} & \textbf{25.23} & \textbf{19.22} & \textbf{13.04}  \\
\midrule
PGD7-AT & 69.85 & 27.78 & 28.44 & 27.87 & 26.00 & 24.75 & 23.86 & 22.8  \\
\textbf{SegPGD7-AT} & 70.21 & \textbf{29.59} & \textbf{30.68} & \textbf{32.55} & \textbf{27.13} & \textbf{25.56} & \textbf{24.29} & \textbf{23.13}   \\
\midrule
\toprule
DeepLabv3 & Clean & CW & DeepFool & BIMl2 & PGD10 & PGD20 & PGD40 & PGD100   \\
\midrule
Standard & 73.82 & 8.24 & 14.26 & 13.86 & 1.07 & 0.84 & 0.62 & 0.44   \\
\midrule
DDCAT~\cite{xu2021dynamic} & 71.68  & \textbf{50.63} & \textbf{54.10} & \textbf{47.42} & 24.45 & 20.80 & \textbf{17.88} & 14.82  \\
PGD3-AT & 71.45 & 36.72 & 38.98 & 36.78 & 29.52 & 20.23 & 12.22 & 6.74  \\
\textbf{SegPGD3-AT} & 71.04 & 37.93 & 37.63 & 34.54 & \textbf{32.11} & \textbf{25.49} & 17.67 & \textbf{15.23}  \\
\midrule
PGD7-AT & 69.91 & 28.87 & 29.63 & 30.58 & 25.64 & 24.48 & 22.87 & 21.24  \\
\textbf{SegPGD7-AT} & 69.93 & \textbf{29.73} & \textbf{31.30} & \textbf{32.35} & \textbf{30.43} & \textbf{28.78} & \textbf{26.73} & \textbf{25.31} \\
\bottomrule
\end{tabular} \vspace{-0.4cm}
\end{center}
\end{table}

\vspace{0.1cm}
\noindent\textbf{Ablation on Weighting Schedules.} In this work, we argue that the weight should be changed dynamically with the attack iterations. At the beginning of the attack, the update of adversarial example should focus more on fooling correctly classified pixels. In Equation~\ref{equ:lambda}, we list three schedule instances, i.e., the \textit{weight-dyn-linear},  \textit{weight-dyn-exp}, and \textit{weight-dyn-log} schedule respectively. We denote the case as \textit{baseline} where the losses of all the pixels are equally treated. Another choice to weigh the loss of misclassified pixels is to use a constant $\lambda$, \textit{e.g.,} 0.1, 0.2 or 0.3, which is denoted as \textit{weight-mis-$\lambda$}. When the constant is set to zero, only correctly classified pixels are considered to compute the loss in all attack iterations, which is denoted as \textit{only-correct}. We report the mIoU of segmentation under different weighting schedules in Fig.~\ref{fig:weight_schedues}. As shown in the figure, SegPGD with our \textit{weight-dyn-linear} weighting schedules can better reduce the mIoU and achieve better attack effectiveness than baselines. Given its simplicity, we apply the linear schedule rule in our SegPGD. We leave the exploration of more dedicated weighting schedules in future work.

\subsection{Boosting Segmentation Robustness with SegPGD-AT}

The setting of adversarial training in previous work~\cite{xu2021dynamic} is adopted in this work. In the baseline, PGD is applied as the underlying attack method for adversarial training. In our approach,
We apply SegPGD to create adversarial examples for adversarial training. For both standard training and adversarial training, we train for one more time and report the average results.

\vspace{0.1cm}
\noindent\textbf{White-Box Attack.} We evaluate the segmentation models with popular white-box attacks. The results are reported in Tab.~\ref{tab:seg_base_voc} and Tab.~\ref{tab:seg_base_city}. The mIoU of the standard segmentation model can be reduced to near zero. As expected, they are not robust at all to strong attack methods. Adversarial training methods boost the robustness of segmentation models to different degrees. Under the evaluation of all attack methods, adversarial training with our SegPGD achieves more robust segmentation performance than the one with PGD. Besides the popular segmentation attack methods, we also evaluate the adversarially-trained models with our SegPGD. The evaluation results also support our conclusion, which can be found in Appendix C.

We also compare our SegPGD-AT with the recently proposed segmentation adversarial training method DDCAT~\cite{xu2021dynamic}. We load the pre-trained DDCAT models from their released the codebase and evaluate the model with strong attacks. We found that their models are vulnerable to strong attacks, \textit{e.g.,} PGD100. For fair comparison, we compare the scores on our SegPGD3-AT with the ones on their models since three steps are applied to generate adversarial examples in both cases. Our model trained with SegPGD3-AT outperform the DDCAT by a large margin under strong attacks, \textit{e.g.,} 10.98 (DDCAT) vs. 18.24 (ours) with DeepLabv3 architecture on VOC dataset under PGD100. More results can be found in Appendix D.

In our experiments, PGD-AT PSPNet on Cityscapes can be almost completely fooled under strong attack where the mIoU is 3.95 under PGD100 attack. Adversarial training with SegPGD boosts the robustness to 13.04. Although the improvement is large, there is still much space to improve.

\vspace{0.1cm}
\noindent\textbf{Black-Box attack.} We also evaluate the segmentation robustness with black-box attacks. Different from white-box attacks, black-box attackers are supposed to have no access to the gradient of the target model. Following the previous work~\cite{xu2021dynamic}, we conduct experiments with transfer-based black-box attacks. We train PSPNet and DeepLabV3 on the same dataset. Then, we create adversarial examples on PSPNet with PGD100 or SegPGD100 and test the robustness of DeepLabV3 on these adversarial examples. The detailed results are reported in Appendix E. The DeepLabV3 models trained with different adversarial training methods are tested. The model trained with our SegPGD-AT shows the best performance against the transfer-based black-box attacks. The claim is also true when different attack methods are applied to create adversarial examples. 

\section{Conclusions}
A large number of attack iterations are required to create effective segmentation adversarial examples. The requirement makes both robustness evaluation and adversarial training on segmentation challenging. In this work, we propose an effective and efficient segmentation-specific attack method, dubbed SegPGD. We first show SegPGD can converge better and faster than the baseline PGD. The effectiveness and efficiency of SegPGD is verified with comprehensive experiments on different segmentation architectures and popular datasets. Besides the evaluation, we also demonstrate how to boost the robustness of segmentation models with SegPGD. Specifically, we apply SegPGD to create segmentation adversarial examples for adversarial training. Given the high effectiveness of the created adversarial examples, the adversarial training with SegPGD improves the segmentation robustness significantly and achieves the state of the art. However, there is still much space to improve in terms of the effectiveness and efficiency of segmentation adversarial training. We hope this work can serve as a solid baseline and inspire more work to improve segmentation robustness.

\vspace{0.1cm}
\noindent\textbf{Acknowledgement} This work is supported by the UKRI grant: Turing AI Fellowship EP/W002981/1, EPSRC/MURI grant: EP/N019474/1, HKU Startup Fund, and HKU Seed Fund for Basic Research. We would also like to thank the Royal Academy of Engineering and FiveAI.
\bibliographystyle{splncs04}
\bibliography{egbib}

\clearpage
\appendix

\vspace{1cm}
\section*{\centering \Large SegPGD: An Effective and Efficient Adversarial Attack for Evaluating and Boosting Segmentation Robustness} 

\vspace{0.5cm}
\section*{\centering \large Supplementary Material}

\vspace{1cm}
\section{\normalfont Comparison of SegPGD with other Segmentation Methods}
\begin{figure}[!h]
    \centering
    \begin{subfigure}[b]{0.48\textwidth}
    \centering
    \includegraphics[scale=0.58]{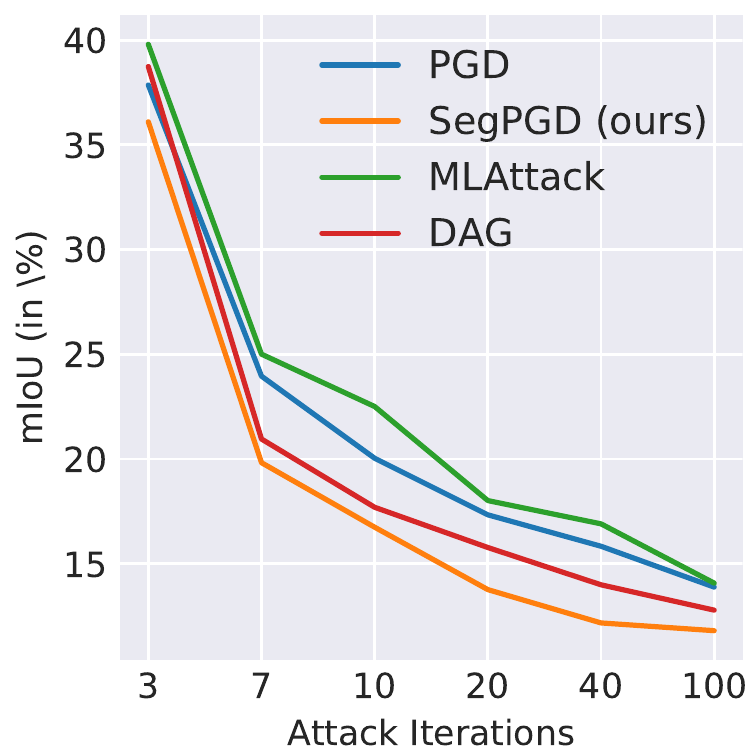}\vspace{-0.1cm} 
    \caption{\scriptsize PSPNet trained with PGD3-AT on VOC}
    \end{subfigure} \hspace{0.05cm} 
    \begin{subfigure}[b]{0.48\textwidth}
    \centering
    \includegraphics[scale=0.58]{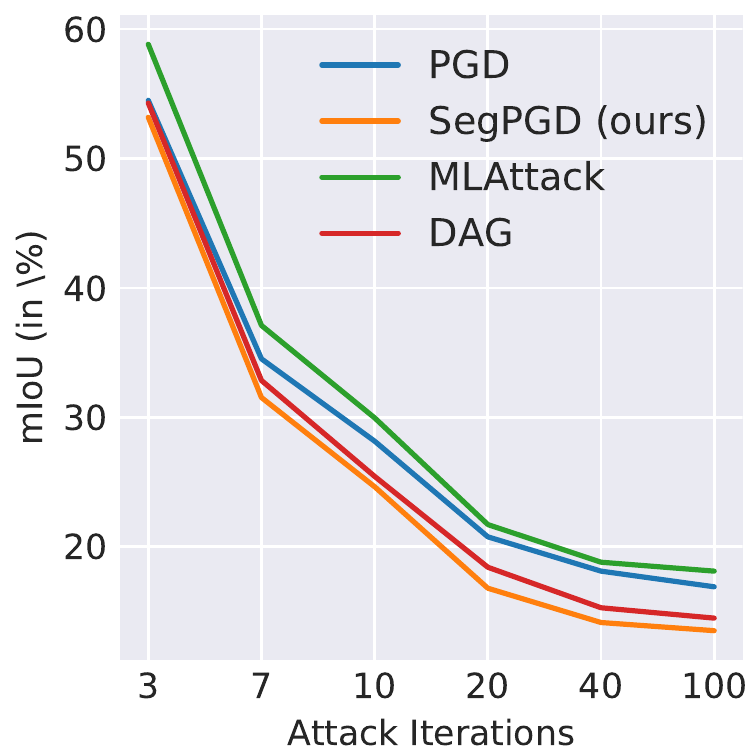}\vspace{-0.1cm} 
    \caption{\scriptsize DeepLabV3 trained with PGD3-AT on VOC}
    \end{subfigure}\vspace{0.2cm}
    \caption{Comparison of SegPGD with other Segmentation Methods. Given the same computational cost (\textit{i.e.,}, the same number of propagation passes), our SegPGD achieves better attack effectiveness.}
    \label{fig:com_other_seg}
\end{figure}

We report the robust accuracy of adversarially trained (PGD3-AT) models under different attacks, namely, SegPGD, DAG and MLAttack. In DAG method, we apply projected gradient descent as the underlying optimization method and only focus on the correctly classified pixels. In MLAttack, three losses are considered for each input image, \textit{i.e.,} the segmentation loss in the output layer, the of in the last layer of encoder and the MSE loss of features multiple Note that the MSE loss is computed as the MSE between the features on the clean input and the ones on current adversarial examples. For each of the three losses, the input gradients are computed to update the input examples. For fair comparison, we compare the segmentation methods with the same number of gradient propagation passes. As shown in Fig.~\ref{fig:com_other_seg}, our SegPGD achieves better attack effectiveness and converges faster than other segmentation methods.

\vspace{0.5cm}

\vspace{0.5cm}
\section{\normalfont Single-step Attack: SegFGSM}
When a single-step attack iteration is applied, SegPGD is degraded to SegFGSM. The results under the single-step attack is shown in Tab.~\ref{tab:seg_fgsm}. As shown in the table, our SegFGSM  outperforms  FGSM  on  both  standard models and adversarially trained models. The conclusion is true across popular segmentation model architectures on two standard segmentation datasets.

\begin{table}[!h]
\begin{center}
\scriptsize
\setlength\tabcolsep{0.12cm}
\begin{tabular}{c | cc | cc |cc |cc}
\toprule
& \multicolumn{2}{|c|}{PSPNet-VOC} & \multicolumn{2}{|c|}{DeepLabV3-VOC} & \multicolumn{2}{|c}{PSPNet-CityScapes} & \multicolumn{2}{|c}{DeepLabV3-CityScapes}  \\
\midrule
& Standard & AT &  Standard & AT &  Standard & AT & Standard & AT \\
\midrule
Clean & 76.64 & 74.51 & 77.36 & 75.03 & 73.98 & 71.28 & 73.82 & 71.45 \\
\midrule
FGSM & 36.76 & 55.33  & 37.59 & 46.78 & 43.76 & 57.5 & 42.79 & 53.85 \\
\midrule
\textbf{SegFGSM} & \textbf{30.80} & \textbf{53.98} & \textbf{31.58} & \textbf{43.88} & \textbf{38.53} & \textbf{56.53} & \textbf{37.97} & \textbf{52.92} \\
\bottomrule
\end{tabular}
\end{center}
\caption{Single-step Attack. Our SegFGSM ourperforms FGSM on both standard models and adversarially trained models.}
\label{tab:seg_fgsm}
\end{table}

\section{\normalfont Model Evaluation under SegPGD Attack}
We evaluate adversarial trained SegPGD-AT models with our SegPGD attack method. As shown in Tab.~\ref{tab:segpgd_eval}, the model adversarially trained with SegPGD also outperforms the one with PGD under the SegPGD attack evaluation. In addition, the observation also echos our claim that the SegPGD can better fool segmentation models than PGD.

\begin{table}[!h]
\begin{center}
\scriptsize
\setlength\tabcolsep{0.18cm}
\begin{tabular}{c | c | cc | cc}
\toprule
\multicolumn{2}{c}{} &  \multicolumn{4}{|c}{AT on VOC}   \\
\cmidrule{3-6}
\multicolumn{2}{c|}{} & PGD3-AT & SegPGD3-AT &  PGD7-AT & SegPGD7-AT  \\
\midrule
\multirow{2.5}{*}{Attack Method} &  PGD100 & 13.89 & 14.49 & 16.97 & 19.23  \\
\cmidrule{2-6}
&  SegPGD100 & 9.67  & 10.34 & 16.20 & 17.03   \\
\midrule
\toprule
\multicolumn{2}{c}{}  & \multicolumn{4}{|c}{AT on Cityscapes}  \\
\cmidrule{3-6}
\multicolumn{2}{c|}{} & PGD3-AT & SegPGD3-AT &  PGD7-AT & SegPGD7-AT  \\
\midrule
\multirow{2.5}{*}{Attack Method} &  PGD100 & 3.95 & 13.04 & 22.80 & 23.13 \\
\cmidrule{2-6}
&  SegPGD100 & 1.91 &  8.86 &  17.03 & 22.54 \\
\bottomrule
\end{tabular}
\end{center}
\caption{Model Evaluation under SegPGD Attack. The evaluation on SegPGD-AT PSPNet is reported with mIoU metric.}
\label{tab:segpgd_eval}
\end{table}

\section{\normalfont Comparison of SegPGD-AT with DDCAT}
We also compare our SegPGD-AT with the recently proposed segmentation adversarial training method DDCAT. We load the pre-trained DDCAT models from their released the codebase and evaluate the model with strong attacks. We found that their models are very weak to defend strong attacks. For fair comparison, we compare the scores on our SegPGD3-AT with the ones on their models since three steps are applied to generate adversarial examples in both case. As shown in Tab.~\ref{tab:comp_ddcat}, our model trained with SegPGD3-AT outperform the DDCAT by a large margin under strong attacks.
\begin{table}[!h]
\begin{center}
\scriptsize
\setlength\tabcolsep{0.18cm}
\begin{tabular}{c | c | ccc |cccccc}
\toprule
\multicolumn{2}{c}{} &  \multicolumn{3}{|c}{Attack on PSPNet}  &  \multicolumn{3}{|c}{Attack on DeepLabV3}   \\
\cmidrule{3-8}
\multicolumn{2}{c|}{} & PGD20 & PGD40 &  PGD100 & PGD20 & PGD40 &  PGD100  \\
\midrule
\multirow{2.5}{*}{AT-Models} &  DDCAT~\cite{xu2021dynamic} & 18.96 & 14.22 & 10.84 & 15.23 & 11.27 & 10.98    \\
\cmidrule{2-8}
 & SegPGD3-AT & 20.69 & 17.19 & 14.49 & 20.92 & 19.10 & 18.24\\
\bottomrule
\end{tabular}
\end{center}
\caption{Comparison of SegPGD-AT with DDCAT. The SegPGD-AT model shows higher robust accuracy than DDCAT model under the same attack.}
\label{tab:comp_ddcat}
\end{table}\vspace{-0.1cm}

\vspace{-0.5cm}
\section{\normalfont Black-box Attack on Adversarially Trained Models}
We train PSPNet and DeepLabV3 on the same dataset. Then, we create adversarial examples on PSPNet with PGD100 or SegPGD100 and test the robustness of DeepLabV3 on these adversarial examples. The results are reported in Tab.~\ref{tab:black_box}. We test the DeepLabV3 models trained with different methods. The model trained with our SegPGD3 shows the best performance against the transfer-based black-box attacks. The claim is also true when different attack methods are applied to create adversarial examples.
\begin{table}[!h]
\begin{center}
\scriptsize
\setlength\tabcolsep{0.2cm}
\begin{tabular}{c | c | c | c | c | c }
\toprule
\multicolumn{1}{c}{} & \multicolumn{1}{c}{} & \multicolumn{1}{c}{} & \multicolumn{3}{|c}{Target Model: Deeplabv3 on VOC}  \\
\cmidrule{1-6}
\multirow{3}{*}{\thead{\scriptsize Source Model:\\ \scriptsize PSPNet}} & Training & Attack & PGD3-AT & DDCAT & \textbf{SegPGD3-AT}   \\
\cmidrule{2-6}
& \multirow{2}{*}{PGD3-AT} & PGD100 & 15.98 & 14.87 & \textbf{16.94}    \\
&   & SegPGD100 & 12.38 & 11.94 & \textbf{13.43}   \\
\midrule
\midrule
\multicolumn{1}{c}{} & \multicolumn{1}{c}{} & \multicolumn{1}{c}{} & \multicolumn{3}{|c}{Target Model: Deeplabv3 on Cityscapes}  \\
\cmidrule{1-6}
\multirow{3}{*}{\thead{\scriptsize Source Model:\\ \scriptsize PSPNet}} & Training & Attack & PGD3-AT & DDCAT & \textbf{SegPGD3-AT}   \\
\cmidrule{2-6}
& \multirow{2}{*}{PGD3-AT} & PGD100 & 14.28 & 15.02 & \textbf{19.42}    \\
&   & SegPGD100 & 13.32 & 14.26 & \textbf{20.11}   \\
\bottomrule
\end{tabular}
\end{center}
\caption{Evaluation under Black-box Attacks. The model with our SegPGD3-based adversarial training performs more robust than other methods on different datasets under different attacks.}
\label{tab:black_box}
\end{table}
\end{document}